\pdfoutput=1

\documentclass[11pt]{article}

\usepackage{emnlp2021}

\usepackage{times}
\usepackage{latexsym}

\usepackage[T1]{fontenc}

\usepackage[utf8]{inputenc}

\usepackage{microtype}

\usepackage{graphicx}

\title{On the Multilingual Capabilities of Very Large-Scale English Language Models}

\author{Jordi Armengol-Estap\'e, Ona de Gibert Bonet, and Maite Melero\\
Text Mining Unit \\
Barcelona Supercomputing Center\\
\texttt{\{jordi.armengol,ona.degibert,maite.melero\}@bsc.es} \\
}

\begin{document}
\maketitle
\begin{abstract}
Generative Pre-trained Transformers (GPTs) have recently been scaled to unprecedented sizes in the history of machine learning. These models, solely trained on the language modeling objective, have been shown to exhibit outstanding few-shot learning capabilities in a number of different tasks. Nevertheless, aside from anecdotal experiences, little is known regarding their multilingual capabilities, given the fact that the pre-training corpus is almost entirely composed of English text. In this work, we investigate the multilingual skills of GPT-3, focusing on one language that barely appears in the pre-training corpus, Catalan, which makes the results especially meaningful; we assume that our results may be relevant for other languages as well. We find that the model shows an outstanding performance, particularly in generative tasks, with predictable limitations mostly in language understanding tasks but still with remarkable results given the zero-shot scenario. We investigate its potential and limits in extractive question-answering and natural language generation, as well as the effect of scale in terms of model size.
\end{abstract}


\section{Introduction}

Improving Natural Language Understanding (NLU) and Generation (NLG) by pre-training autoregressive language models based on the Transformer \cite{DBLP:journals/corr/VaswaniSPUJGKP17} decoder architecture has been commonplace since the original GPT (Generative Pretrained Transformer) \cite{Radford2018ImprovingLU} first appeared. In the race to scale up these language models \cite{radford2019language}, the arrival of GPT-3  \cite{DBLP:journals/corr/abs-2005-14165} has changed the rules of the game. As claimed by their creators, its ability to learn from a few examples "via text interaction" makes it stand out from the rest. Its impressive generative capabilities have caused a big sensation, not only at research level but also in the mainstream media. 

A particular feature of GPT-3 is, besides the sheer size of the data it has been trained on, the fact that, although the data is generally of good quality, it has not been filtered for language (in purpose). Therefore, although GPT-3 is in principle a language model for English, its training data contains many other languages,\footnote{\url{https://github.com/openai/gpt-3/tree/master/dataset_statistics}} even if they account for a small portion of the dataset in comparison to English (93\% by word count). Intuitively, one would expect that this quantity would not be enough to obtain a high-quality language model in these other languages, especially in the low-resource ones. Some evidence in this regard is provided by the large amount of data required to train language-specific models \cite{DBLP:journals/corr/abs-2003-02912}. Even the multilingual ones\footnote{Note that both mBERT and XLM-R are encoder-based models, unlike GPT, but the point still holds.} such as mBERT \cite{DBLP:journals/corr/abs-1810-04805} or XLM-R \cite{DBLP:journals/corr/abs-1911-02116} employ large multilingual datasets based on Wikipedia or CommonCrawl. A very recent work trained a language-specific Catalan model with around 1.7B tokens \cite{armengol-estape-etal-2021-multilingual}, but it was published after the elaboration of this article and thus is not included in our comparisons. The code for reproducing the GPT-3 API queries and the results we obtained is openly available.\footnote{\url{https://github.com/TeMU-BSC/gpt3-queries}}

\section{Related Work}

In \citet{DBLP:journals/corr/abs-2005-14165}, the authors of GPT-3 already conducted a thorough evaluation in many different benchmarks, including question-answering, cloze tasks, and Natural Language Inference (NLI), among many others. Crucially, they train and evaluate models of different sizes, and find that by simply scaling up the exact same architecture, the diminishing returns that one would expect are not observed. Recently, some works have estimated the increase in performance of autoregressive models in terms of model size, data, and compute \cite{DBLP:journals/corr/abs-2001-08361, DBLP:journals/corr/abs-2010-14701}. Also in \citet{DBLP:journals/corr/abs-2005-14165}, and relevant to our work, authors evaluate GPT-3 in machine translation, both in zero and few-shot settings, and find that in the latter, GPT-3 outperforms previous unsupervised NMT models by 5 BLEU in some pairs. Specifically, this success is observed in the evaluated pairs in which English is the target language and not in the ones in which English is the source one, being GPT-3 an English language model. No other analysis involving languages other than English was conducted.

Since the original article of GPT-3, several works have investigated the capabilities and limits of the model in English \cite{zhao2021calibrate}. Moreover, with the possibility of querying the model via API, hundreds of researchers, journalists and curious alike have embarked on all sorts of experiments, including automatic programming or solving arithmetic operations \cite{floridi2020gpt}. The Internet is full of examples of the amazing generative capabilities of the model, from poetry, news or essay writing \cite{elkins2020can}.

Furthermore, many researchers are interested in the ethical concerns regarding such a capable generative model and studying the impact it may had if it was released to the public \cite{dale2021gpt,mcguffie2020radicalization}. In a more consequential approach, with the purpose of harnessing the full learning potential of GPT, we are seeing the emergence of a new line of research exploring optimal ways to "prompt" the model \cite{DBLP:journals/corr/abs-2101-06804}.

Nevertheless, to our knowledge, no work has studied its potential for solving tasks in languages other than English, aside from machine translation. In this work, we investigate the multilingual skills of GPT-3, focusing on Catalan, a language barely appearing in the pre-training corpus.

\section{Methodology}
In this work we have explored how good GPT-3 is at generating natural text in Catalan and solving one NLU task, specifically extractive Q\&A. Catalan only accounts for the 0,01798\% of words in the training corpus, that is around 35M words. Language models, even if in a considerably smaller scale than GPT-3, are usually trained on corpora with a number of tokens in the billions as can be seen in Table \ref{tab:languages-words}.  Even considering the effect of certain factors particular to each language, such as linguistic proximity to English (e.g. being an Indo European language), affiliation to well-populated families (e.g. Romance), number of tokens in the training corpus, etc. we can assume that our results may be relevant for other languages as well.

\begin{table}[]
\centering
\begin{tabular}{l|r|r}
\textbf{Model} & \textbf{Words (M)} & \textbf{Catalan words (M)}\\ \hline
mBERT & Unclear\footnotemark & \textasciitilde 200 \\
XLM-R & 295,008\footnotemark & 1,752\\
GPT-3 & 196,755\footnotemark & \textbf{35}\\
\end{tabular}
\caption{\label{tab:languages-words} Pre-training word count in some models}
\end{table}
\footnotetext[3]{mBERT was trained with the top 100 largest Wikipedias, but there are no details on the exact amount of tokens. For Catalan, we estimate the size in 200M tokens from a dump from January 2020.}
\footnotetext[4]{Summing up tokens from all languages from Table 6 in \citet{DBLP:journals/corr/abs-1911-02116}.}
\footnotetext[5]{In the dataset statistics in Github, OpenAI claims that English, with around 181B tokens, accounts for about 93\% of the dataset. This implies a total size of around 197B tokens, the one we use in the table. However, in the article authors say the model was trained with a total of 300B tokens. We have not been able to clarify this apparent inconsistency.}



\subsection{Question-answering}
To evaluate GPT-3 in question-answering, we use a Catalan translation (introduced in \citet{armengol-estape-etal-2021-multilingual}, \citet{carlos_gerardo_rodriguez_penagos_2021_4526224}) of XQuAD \cite{artetxe2019cross}, a cross-lingual question-answering dataset consisting of 240 paragraphs and 1,060 question-answer pairs. We focus on the zero-shot setting, in which the model is not given any example. GPT-3 is asked to answer one question at a time, pieced with its context as prompts as shown below (in bold, GPT-3's answer):
\begin{quote}
\em
Això és un sistema de resposta de preguntes en català.

Context: La defensa dels Panthers va cedir només 308 punts [...]

Pregunta: Quants punts va cedir la defensa dels Panthers?

Resposta:
\textbf{308 punts}
\end{quote}

The whole prompt, including the instruction to answer the question (the first sentence), the context, the question (\textit{Pregunta}), and the final word (\textit{Resposta}, "Answer") are given in Catalan, with the hope that this will further condition the model to answer in Catalan. To study the effect of scale, we run the model with the 4 engines provided in OpenAI's API,\footnote{\url{https://beta.openai.com/}} in increasing size\footnote{To the best of our knowledge, OpenAI has not clarified the exact size of each of the models in the API. However, some evaluations results seem to suggest that Ada, Babbage, Curie and Davinci would correspond to 350M, 1.3B, 6.7B, and 175B, respectively. See: \url{https://blog.eleuther.ai/gpt3-model-sizes/}. } (in parameters): Ada, Babbage, Curie, and Davinci, using the default sampling parameters\footnote{A temperature of 0.7, a frequency penalty of 0, a presence penalty of 0, and with \texttt{top\_p} = 1.} except for \texttt{max\_tokens}, which we set to 64 to allow the longest answers. 

As a reference, we include the results of what should be considered state-of-the-art, the ones obtained by fine-tuning mBERT and XLM-RoBERTa (\textit{base} size for both models) in a Catalan question-answering dataset \cite{rodriguez_penagos_carlos_gerardo_2021_4562345} using the script from the Huggingface library \cite{DBLP:journals/corr/abs-1910-03771} used for fine-tuning on the SQuAD dataset. For all models (including GPT-3), we apply the same evaluation script as in SQuAD.\footnote{\url{https://github.com/allenai/bi-att-flow/blob/master/squad/evaluate-v1.1.py}}

\subsection{Natural Language Generation}
In order to evaluate the generative capabilities of GPT-3 in Catalan, we want to assess how “natural” the generated text is to Catalan natives. For this, we create a synthetic set of 60 sentences and mix them randomly with 60 control sentences coming from a news corpus,\footnote{2021 crawling from \url{https://www.acn.cat/} in Catalan} and ask our evaluators to score each sentence based on their overall fluency and correctness.
To obtain the synthetic sentences, we first query GPT-3 with a set of 20 headlines extracted from the same news corpus, and then sample 60 sentences from the generated output. For this evaluation we only use the output of the largest version of GPT-3 (i.e. Davinci).
We manually checked that the sentences did not appear in the Internet,\footnote{By searching them on Google. None of the sentences appeared verbatim although we removed  a similar one.} to avoid sentences that could have been directly memorized in training. As in question-answering, we used the default sampling parameters of OpenAI's API, this time, setting \texttt{max\_tokens} to 1024, for generating more sentences to sample from.
For the human evaluation, similarly to \cite{casas2020syntax}, sentences were evaluated by a pool of 9 annotators, who were requested to rate the sentence in an integer scale from 1 to 5. Each sentence, randomly distributed among the pool of evaluators, was scored by 3 different evaluators; this redundancy accounts for the variance and subjectivity in human scores.


\begin{table}[]
\centering
\begin{tabular}{l|c|c}
\textbf{Model} & \textbf{F1} & \textbf{EM} \\ \hline
GPT-3: Ada  & 5.26       & 0.38  \\
GPT-3: Babbage  & 10.08       & 1.13\\
GPT-3: Curie  & 16.66      & 5.00\\
GPT-3: Davinci  & \textbf{38.43}       & \textbf{17.74}\\
\hline
XLM-RoBERTa & 67.10 & 46.42  \\
mBERT  & \textbf{67.15}       & \textbf{46.51}\\

\end{tabular}
\caption{\label{tab:qa} Question answering results for XQuAD-ca}
\end{table}

\section{Results}
\paragraph{Question-answering}

\begin{figure}
    \centering
    \includegraphics[scale=0.5]{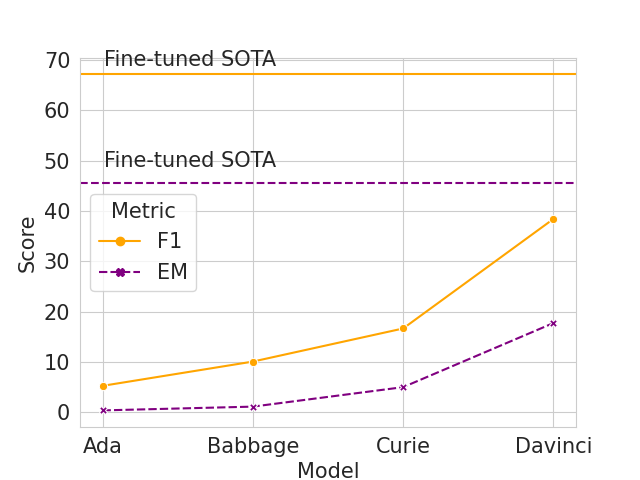}
    \caption{Question-answering results for GPT-3 sizes}
    \label{fig:qa-scores}
\end{figure}

The results obtained by GPT-3 in this task are reported in table \ref{tab:qa}, showing the F1 score and the Exact Match value for XQuAD-ca, for the different GPT-3 model sizes. We also include the results of two supervised, fine-tuned models considered state-of-the art as a reference. Note that this is not a direct comparison, since for GPT-3 it is a zero-shot setting. GPT-3 Davinci obtains a F1 score that is more than 50\% the punctuation obtained by the SOTA models, which is remarkable being a pure zero-shot setting. Figure \ref{fig:qa-scores} shows the scaling curves of the different model-sizes of GPT-3. 


\begin{figure}
    \centering
    \includegraphics[scale=0.5]{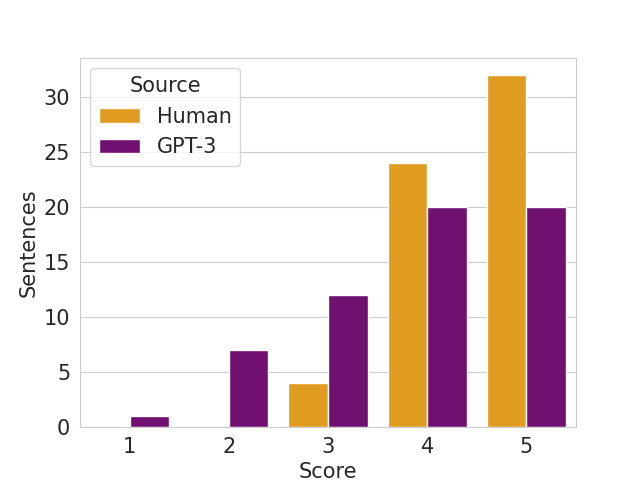}
    \caption{Distribution of Human Evaluation ratings}
    \label{fig:human-eval}
\end{figure}

\paragraph{Natural Language Generation}
Table \ref{tab:human_eval_stats} shows the results of the human evaluation. The sentences generated by GPT-3 obtain an average score of 3,89, compared to 4,49 of the control.\footnote{The difference is statistically significant. With a t-test, we obtain a p-value of 0.00026 < 0.001.} As can be seen by the difference between the standard deviations and the distribution of scores in Figure \ref{fig:human-eval}, GPT-3 is less consistent than the control in quality, however most of the sentences are rated between 4 and 5 by the evaluators. In fact, a third of the sentences is above the average of the control, versus half of the ones generated by humans.



\begin{table}[]
\centering
\begin{tabular}{l|c|c|c}
\textbf{Source} & \textbf{\begin{tabular}{c}Average \\Rating\end{tabular}}& \textbf{St. Dev.} & \textbf{\begin{tabular}{c} \% >\\Human Av. \end{tabular}} \\ 
\hline
Human & 4.49 & 0.57 & 53.33\\
GPT-3 & 3.83 & 1.05 & 33.33 \\     
\end{tabular}
\caption{\label{tab:human_eval_stats} Human evaluation (for GPT-3, Davinci)}
\end{table}

\section{Discussion}

\paragraph{Qualitative analysis}
A closer inspection of the results shows some surprising abilities of GPT-3 in addition to the naturalness of most of the sentences. An interesting example is that following the prompt of a headline about Valencia,  GPT-3 is able to write using the Valencian variant of Catalan, which is truly remarkable. An analysis of the errors shows that those with score of 2 or less (13\% of the sample) contain gibberish fragments, often mixing Catalan and English, and in fact no control sentence has received such low scores. On the other hand, sentences with score 3 (21,6\%) are mostly syntactically impeccable but with some peculiarities in the meaning, as for example: 
"La IV Mostra de Patrimoni Cultural de Bétera ha comptat amb una participació de 15.000 persones, que han pogut gaudir d'un espai on diversos grups han mostrat \textit{els seus valors patrimonials}."



\paragraph{Scaling} As shown in Figure \ref{fig:qa-scores}, there is a steep curve of F1 score in terms of model size, while pre-training data (and, thus, the amount of Catalan) remains the same. This shows that transfer learning between English and the other languages in zero-shot settings scales with model size in a very steep curve. This is coherent with Figure H.11 in \citet{DBLP:journals/corr/abs-2005-14165}, where zero-shot translation in which English is the target language reaches a plateau, but when the target languages are languages other than English, the curves keep climbing.

\paragraph{Usability in practice} We believe the model can be useful in multilingual applications (at least, in a degree not far from the one for English), especially since we used the model in zero-shot settings and without any effort in prompt design. We expect the model to perform considerably better in few-shot settings, and even better in languages with more data in GPT-3's corpus. Nevertheless, a caveat, at least for Catalan, is that smaller versions of GPT-3 aren't usable, and because the vocabulary was trained fundamentally on English, Catalan sentences are tokenized into considerably long sequences, which makes them expensive to compute.

\paragraph{Limitations of our study} We have restricted our analysis to the case of Catalan, and to two specific tasks, even if we believe them to be relevant, and reasonably representative of the NLP scenario. We have constrained the analysis to the zero-shot setting, which we believe to be the most interesting one. For the human evaluation, we have tried to make it as balanced as possible by using a redundancy of 3 evaluators, but human ratings can be biased. Regarding the relevance to other languages, as already mentioned, Catalan probably benefits from linguistic similarities with Romance and Indo European languages at large (including English). 


\section{Conclusions and Future Work}
We have seen that GPT-3 does, indeed, exhibit remarkable zero-shot NLU and NLG capabilities in Catalan. This is surprising in view of the tiny proportion of Catalan in the training corpus. Our results show that GPT-3 can be useful not only for English but for many other languages present in the corpus as well. Nevertheless, some practical concerns (the needed model scale and sub optimal tokenization) make it less computationally efficient than for English. 
On the overall, this is a very interesting exercise of how linguistic structures (universals) transfer across languages.
Given the large amount of tasks GPT-3 has been implicitly exposed to during the training procedure, handling a different language can be considered as working on yet another domain. As future work, we suggest extending the study of the scaling laws of language models \cite{DBLP:journals/corr/abs-2001-08361} in terms of cross-lingual transfer, similarly to \citet{DBLP:journals/corr/abs-2102-01293}. 



\bibliography{main}

\begin{thebibliography}{22}
\expandafter\ifx\csname natexlab\endcsname\relax\def\natexlab#1{#1}\fi

\bibitem[{Armengol-Estap{\'e} et~al.(2021)Armengol-Estap{\'e}, Carrino,
  Rodriguez-Penagos, de~Gibert~Bonet, Armentano-Oller, Gonzalez-Agirre, Melero,
  and Villegas}]{armengol-estape-etal-2021-multilingual}
Jordi Armengol-Estap{\'e}, Casimiro~Pio Carrino, Carlos Rodriguez-Penagos, Ona
  de~Gibert~Bonet, Carme Armentano-Oller, Aitor Gonzalez-Agirre, Maite Melero,
  and Marta Villegas. 2021.
\newblock \href {https://doi.org/10.18653/v1/2021.findings-acl.437} {Are
  multilingual models the best choice for moderately under-resourced languages?
  {A} comprehensive assessment for {C}atalan}.
\newblock In \emph{Findings of the Association for Computational Linguistics:
  ACL-IJCNLP 2021}, pages 4933--4946, Online. Association for Computational
  Linguistics.

\bibitem[{Artetxe et~al.(2019)Artetxe, Ruder, and Yogatama}]{artetxe2019cross}
Mikel Artetxe, Sebastian Ruder, and Dani Yogatama. 2019.
\newblock On the cross-lingual transferability of monolingual representations.
\newblock \emph{arXiv preprint arXiv:1910.11856}.

\bibitem[{Brown et~al.(2020)Brown, Mann, Ryder, Subbiah, Kaplan, Dhariwal,
  Neelakantan, Shyam, Sastry, Askell, Agarwal, Herbert{-}Voss, Krueger,
  Henighan, Child, Ramesh, Ziegler, Wu, Winter, Hesse, Chen, Sigler, Litwin,
  Gray, Chess, Clark, Berner, McCandlish, Radford, Sutskever, and
  Amodei}]{DBLP:journals/corr/abs-2005-14165}
Tom~B. Brown, Benjamin Mann, Nick Ryder, Melanie Subbiah, Jared Kaplan,
  Prafulla Dhariwal, Arvind Neelakantan, Pranav Shyam, Girish Sastry, Amanda
  Askell, Sandhini Agarwal, Ariel Herbert{-}Voss, Gretchen Krueger, Tom
  Henighan, Rewon Child, Aditya Ramesh, Daniel~M. Ziegler, Jeffrey Wu, Clemens
  Winter, Christopher Hesse, Mark Chen, Eric Sigler, Mateusz Litwin, Scott
  Gray, Benjamin Chess, Jack Clark, Christopher Berner, Sam McCandlish, Alec
  Radford, Ilya Sutskever, and Dario Amodei. 2020.
\newblock \href {http://arxiv.org/abs/2005.14165} {Language models are few-shot
  learners}.
\newblock \emph{CoRR}, abs/2005.14165.

\bibitem[{Casas et~al.(2020)Casas, Fonollosa, and
  Costa-juss{\`a}}]{casas2020syntax}
Noe Casas, Jos{\'e}~AR Fonollosa, and Marta~R Costa-juss{\`a}. 2020.
\newblock Syntax-driven iterative expansion language models for controllable
  text generation.
\newblock \emph{arXiv preprint arXiv:2004.02211}.

\bibitem[{Conneau et~al.(2019)Conneau, Khandelwal, Goyal, Chaudhary, Wenzek,
  Guzm{\'{a}}n, Grave, Ott, Zettlemoyer, and
  Stoyanov}]{DBLP:journals/corr/abs-1911-02116}
Alexis Conneau, Kartikay Khandelwal, Naman Goyal, Vishrav Chaudhary, Guillaume
  Wenzek, Francisco Guzm{\'{a}}n, Edouard Grave, Myle Ott, Luke Zettlemoyer,
  and Veselin Stoyanov. 2019.
\newblock \href {http://arxiv.org/abs/1911.02116} {Unsupervised cross-lingual
  representation learning at scale}.
\newblock \emph{CoRR}, abs/1911.02116.

\bibitem[{Dale(2021)}]{dale2021gpt}
Robert Dale. 2021.
\newblock Gpt-3: What’s it good for?
\newblock \emph{Natural Language Engineering}, 27(1):113--118.

\bibitem[{Devlin et~al.(2018)Devlin, Chang, Lee, and
  Toutanova}]{DBLP:journals/corr/abs-1810-04805}
Jacob Devlin, Ming{-}Wei Chang, Kenton Lee, and Kristina Toutanova. 2018.
\newblock \href {http://arxiv.org/abs/1810.04805} {{BERT:} pre-training of deep
  bidirectional transformers for language understanding}.
\newblock \emph{CoRR}, abs/1810.04805.

\bibitem[{Elkins and Chun(2020)}]{elkins2020can}
Katherine Elkins and Jon Chun. 2020.
\newblock Can gpt-3 pass a writer’s turing test.
\newblock \emph{Journal of Cultural Analytics}, 2371:4549.

\bibitem[{Floridi and Chiriatti(2020)}]{floridi2020gpt}
Luciano Floridi and Massimo Chiriatti. 2020.
\newblock Gpt-3: Its nature, scope, limits, and consequences.
\newblock \emph{Minds and Machines}, 30(4):681--694.

\bibitem[{Henighan et~al.(2020)Henighan, Kaplan, Katz, Chen, Hesse, Jackson,
  Jun, Brown, Dhariwal, Gray, Hallacy, Mann, Radford, Ramesh, Ryder, Ziegler,
  Schulman, Amodei, and McCandlish}]{DBLP:journals/corr/abs-2010-14701}
Tom Henighan, Jared Kaplan, Mor Katz, Mark Chen, Christopher Hesse, Jacob
  Jackson, Heewoo Jun, Tom~B. Brown, Prafulla Dhariwal, Scott Gray, Chris
  Hallacy, Benjamin Mann, Alec Radford, Aditya Ramesh, Nick Ryder, Daniel~M.
  Ziegler, John Schulman, Dario Amodei, and Sam McCandlish. 2020.
\newblock \href {http://arxiv.org/abs/2010.14701} {Scaling laws for
  autoregressive generative modeling}.
\newblock \emph{CoRR}, abs/2010.14701.

\bibitem[{Hernandez et~al.(2021)Hernandez, Kaplan, Henighan, and
  McCandlish}]{DBLP:journals/corr/abs-2102-01293}
Danny Hernandez, Jared Kaplan, Tom Henighan, and Sam McCandlish. 2021.
\newblock \href {http://arxiv.org/abs/2102.01293} {Scaling laws for transfer}.
\newblock \emph{CoRR}, abs/2102.01293.

\bibitem[{Kaplan et~al.(2020)Kaplan, McCandlish, Henighan, Brown, Chess, Child,
  Gray, Radford, Wu, and Amodei}]{DBLP:journals/corr/abs-2001-08361}
Jared Kaplan, Sam McCandlish, Tom Henighan, Tom~B. Brown, Benjamin Chess, Rewon
  Child, Scott Gray, Alec Radford, Jeffrey Wu, and Dario Amodei. 2020.
\newblock \href {http://arxiv.org/abs/2001.08361} {Scaling laws for neural
  language models}.
\newblock \emph{CoRR}, abs/2001.08361.

\bibitem[{Liu et~al.(2021)Liu, Shen, Zhang, Dolan, Carin, and
  Chen}]{DBLP:journals/corr/abs-2101-06804}
Jiachang Liu, Dinghan Shen, Yizhe Zhang, Bill Dolan, Lawrence Carin, and Weizhu
  Chen. 2021.
\newblock \href {http://arxiv.org/abs/2101.06804} {What makes good in-context
  examples for gpt-3?}
\newblock \emph{CoRR}, abs/2101.06804.

\bibitem[{McGuffie and Newhouse(2020)}]{mcguffie2020radicalization}
Kris McGuffie and Alex Newhouse. 2020.
\newblock The radicalization risks of gpt-3 and advanced neural language
  models.
\newblock \emph{arXiv preprint arXiv:2009.06807}.

\bibitem[{Nozza et~al.(2020)Nozza, Bianchi, and
  Hovy}]{DBLP:journals/corr/abs-2003-02912}
Debora Nozza, Federico Bianchi, and Dirk Hovy. 2020.
\newblock \href {http://arxiv.org/abs/2003.02912} {What the [mask]? making
  sense of language-specific {BERT} models}.
\newblock \emph{CoRR}, abs/2003.02912.

\bibitem[{Radford and Narasimhan(2018)}]{Radford2018ImprovingLU}
A.~Radford and Karthik Narasimhan. 2018.
\newblock Improving language understanding by generative pre-training.

\bibitem[{Radford et~al.(2019)Radford, Wu, Child, Luan, Amodei, and
  Sutskever}]{radford2019language}
Alec Radford, Jeff Wu, Rewon Child, David Luan, Dario Amodei, and Ilya
  Sutskever. 2019.
\newblock Language models are unsupervised multitask learners.

\bibitem[{Rodriguez-Penagos and
  Armentano-Oller(2021{\natexlab{a}})}]{rodriguez_penagos_carlos_gerardo_2021_4562345}
Carlos~Gerardo Rodriguez-Penagos and Carme Armentano-Oller. 2021{\natexlab{a}}.
\newblock \href {https://doi.org/10.5281/zenodo.4562345} {{ViquiQuAD: an
  extractive QA dataset from Catalan Wikipedia}}.

\bibitem[{Rodriguez-Penagos and
  Armentano-Oller(2021{\natexlab{b}})}]{carlos_gerardo_rodriguez_penagos_2021_4526224}
Carlos~Gerardo Rodriguez-Penagos and Carme Armentano-Oller. 2021{\natexlab{b}}.
\newblock \href {https://doi.org/10.5281/zenodo.4526224} {Xquad-ca}.

\bibitem[{Vaswani et~al.(2017)Vaswani, Shazeer, Parmar, Uszkoreit, Jones,
  Gomez, Kaiser, and Polosukhin}]{DBLP:journals/corr/VaswaniSPUJGKP17}
Ashish Vaswani, Noam Shazeer, Niki Parmar, Jakob Uszkoreit, Llion Jones,
  Aidan~N. Gomez, Lukasz Kaiser, and Illia Polosukhin. 2017.
\newblock \href {http://arxiv.org/abs/1706.03762} {Attention is all you need}.
\newblock \emph{CoRR}, abs/1706.03762.

\bibitem[{Wolf et~al.(2019)Wolf, Debut, Sanh, Chaumond, Delangue, Moi, Cistac,
  Rault, Louf, Funtowicz, and Brew}]{DBLP:journals/corr/abs-1910-03771}
Thomas Wolf, Lysandre Debut, Victor Sanh, Julien Chaumond, Clement Delangue,
  Anthony Moi, Pierric Cistac, Tim Rault, R{\'{e}}mi Louf, Morgan Funtowicz,
  and Jamie Brew. 2019.
\newblock \href {http://arxiv.org/abs/1910.03771} {Huggingface's transformers:
  State-of-the-art natural language processing}.
\newblock \emph{CoRR}, abs/1910.03771.

\bibitem[{Zhao et~al.(2021)Zhao, Wallace, Feng, Klein, and
  Singh}]{zhao2021calibrate}
Tony~Z Zhao, Eric Wallace, Shi Feng, Dan Klein, and Sameer Singh. 2021.
\newblock Calibrate before use: Improving few-shot performance of language
  models.
\newblock \emph{arXiv preprint arXiv:2102.09690}.

\end{thebibliography}
\bibliographystyle{acl_natbib}




\end{document}